\documentclass[sigconf]{acmart}
\AtBeginDocument{%
  }

\setcopyright{acmlicensed}
\copyrightyear{2025}
\acmYear{2025}
\acmDOI{10.1145/3772429.3772445}

\acmConference[DAI]{Make sure to enter the correct
  conference title from your rights confirmation email}{NOV 21--24,
  2025}{London}

\usepackage{microtype}
\usepackage[utf8]{inputenc}
\usepackage[T1]{fontenc}
\usepackage{xcolor}
\usepackage{threeparttable}
\usepackage{booktabs}
\usepackage{graphicx}
\usepackage{caption}
\usepackage{nicefrac}
\usepackage{multirow}
\usepackage{enumitem}
\usepackage{url}
\usepackage{array}

\begin{document}
\title{Beyond GPT-5: Making LLMs Cheaper and Better via Performance–Efficiency Optimized Routing}

\author{Yiqun Zhang$^{*,\ddag}$, Hao Li$^*$, Jianhao Chen$^*$, Hangfan Zhang$^*$, Peng Ye, Lei Bai, Shuyue Hu$^{\dag,\ddag}$}
\email{zhangyiqun344@gmail.com, {lihao4,chenjianhao,zhanghangfan,yepeng,bailei,hushuyue}@pjlab.org.cn}
\affiliation{%
  \institution{Shanghai Artificial Intelligence Laboratory}
  \city{Shanghai}
  \country{China}
}

\renewcommand{\shortauthors}{Zhang et al.}
\begin{abstract}
Balancing performance and efficiency is a central challenge in large language model (LLM) advancement. GPT-5 addresses this with test-time routing, dynamically assigning queries to either an efficient or a high-capacity model during inference. In this work, we present \emph{Avengers-Pro}, a test-time routing framework that ensembles LLMs of varying capacities and efficiencies, providing a unified solution for all performance-efficiency tradeoffs.
The \emph{Avengers-Pro} 
embeds and clusters incoming queries, then routes each to the most suitable model based on a performance-efficiency score.
Across 6 challenging benchmarks and 8 leading models---including {GPT-5-medium}, {Gemini-2.5-pro}, and {Claude-opus-4.1}---\emph{Avengers-Pro} achieves state-of-the-art results: by varying a performance-efficiency trade-off parameter, it can \textbf{surpass the strongest single model} ({GPT-5-medium}) by \textbf{+7\% in average accuracy}. Moreover, it can 
\textbf{match} the average accuracy of the strongest single model at  \textbf{27\% lower cost}, and reach $\sim$\textbf{90\%} of that performance at \textbf{63\% lower cost}. Last but not least, it achieves a Pareto frontier, consistently yielding the highest accuracy for any given cost, and the lowest cost for any given accuracy, among all single models.
Code is available at \url{https://github.com/ZhangYiqun018/AvengersPro}.
\def\thefootnote{*}\footnotetext{Equal contribution.}\def\thefootnote{\arabic{footnote}}
\def\thefootnote{\dag}\footnotetext{Corresponding author.}\def\thefootnote{\arabic{footnote}}
\def\thefootnote{\ddag}\footnotetext{Project lead, hushuyue@pjlab.org.cn, zhangyiqun344@gmail.com}\def\thefootnote{\arabic{footnote}}
\end{abstract}



\received{22 August 2025}
\received[accepted]{18 September 2025}

\begin{teaserfigure}
    \centering \includegraphics[width=1\linewidth]{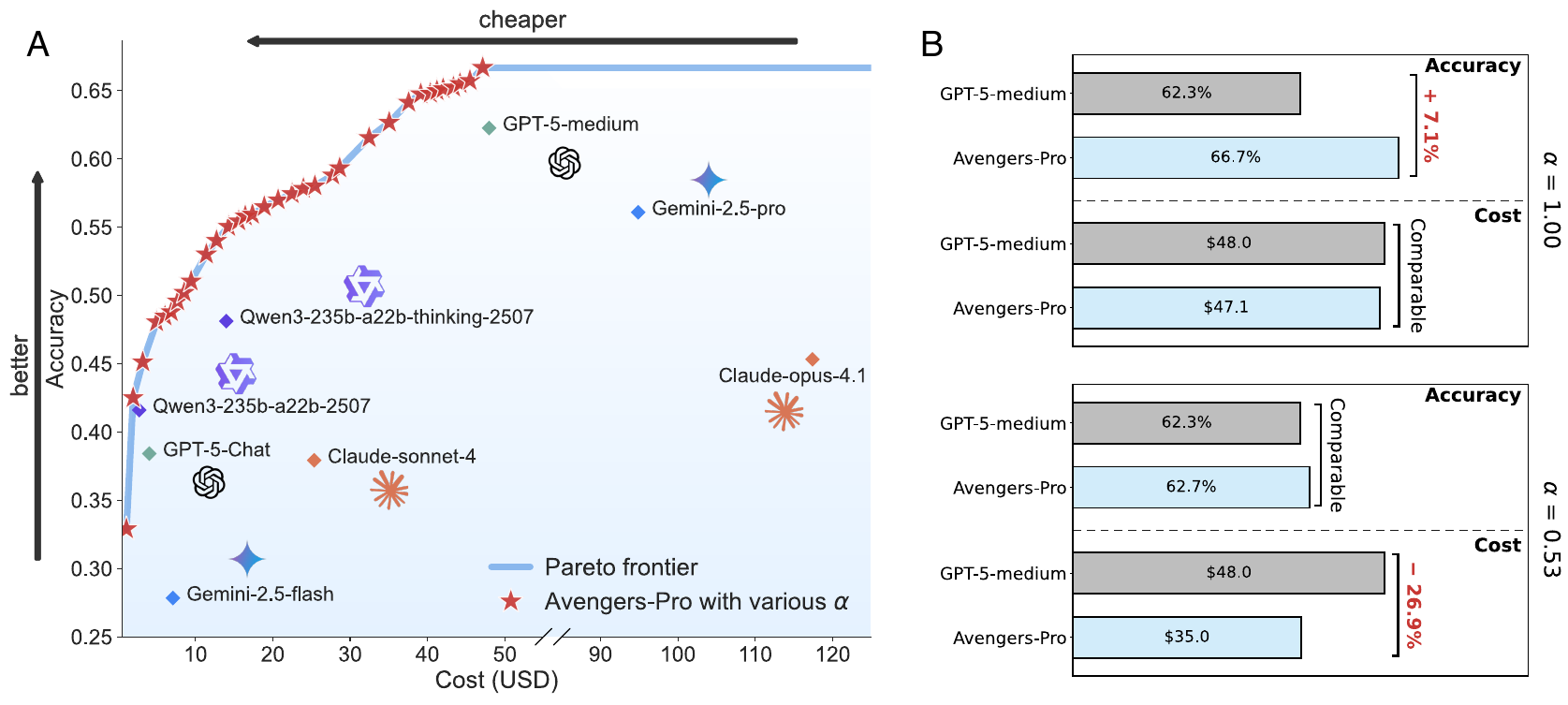}
    \caption{\emph{Avengers-Pro} optimizes the trade-off between performance (accuracy) and efficiency (cost).  \textbf{(A)} By varying a trade-off parameter $\alpha$, \emph{Avengers-Pro} establishes a Pareto frontier. Compared to all single models, it achieves the highest accuracy for any given cost, and achieves the lowest cost for any given accuracy. \textbf{(B)} With comparable cost, \emph{Avengers-Pro} outperforms the strongest single model GPT-5-medium by $7.1\%$. With comparable performance, \emph{Avengers-Pro} achieves a $26.9\%$ cost reduction compared to GPT-5-medium.}
    \label{fig:big-picture}
\end{teaserfigure}

\maketitle

\section{Introduction}
A fundamental dilemma in LLM advancement is the trade-off between performance and efficiency.
To navigate this, 
a defining feature of GPT-5 is its \textit{test-time routing} between models. 
 As described in \textit{Introducing GPT-5}\footnote{https://openai.com/index/introducing-gpt-5/}:
\begin{quote}
``GPT‑5 is a unified system with a \textbf{smart, efficient} model that answers most questions, a \textbf{deeper reasoning} model (GPT‑5 thinking) for harder problems, and \textbf{a real‑time router} that quickly decides which to use based on conversation type, complexity ... ''
\end{quote}
The efficient model offers lower computational cost and latency at the expense of capability, while the deeper reasoning model incurs higher cost and latency but delivers greater capability. During inference, GPT-5’s router dynamically assigns each query to exactly one model,  striking a balance between performance and efficiency.


In this work, we advance test-time routing to optimize the performance–efficiency trade-off, and introduce the \textit{Avengers-Pro}. Given a set of models and  a set of labeled query-answer pairs, the \textit{Avengers-Pro} 
operates through three lightweight operations: embedding, clustering and scoring. 
Specifically, first, it encodes the queries from the  dataset sing a text embedding model, and then clusters them based on their semantic representations. Next, to assess each model’s capabilities and efficiency, it evaluates each model on the dataset and computes a performance-efficiency score for each cluster. Weighted by a trade-off parameter $\alpha$, this score reflects both a model’s performance (measured by its accuracy on the queries within a cluster) and its efficiency (quantified by the cost incurred when answering those queries).
During inference, each query is embedded and mapped to its top-$p$ nearest clusters. The model with the highest performance-efficiency score aggregated over those clusters is selected to generate the response.

In our experiments, the \textit{Avengers-Pro} consists of 8 models from 4 families: {GPT-5-chat}, {GPT-5-medium}, {Claude-4.1-opus}, {Claude-4-sonnet}, {Gemini-2.5-pro}, {Gemini-2.5-flash}, {Qwen3-235B-A22B-thinking-2507}, and {Qwen3-235B-A22B-2507}. We evaluated \textit{Avengers--Pro} on 6 challenging benchmarks: GPQA-Diamond~\cite{rein2024gpqa}, Human's Last Exam~\cite{phan2025humanity}, HealthBench~\cite{arora2025healthbench}, ARC-AGI~\cite{chollet2024arc}, SimpleQA~\cite{wei2024measuring}, LiveCodeBench~\cite{jain2024livecodebench}, and $\tau$2-bench~\cite{barres2025tau}. 
We find that compared to the strongest single model {GPT-5-medium} (average accuracy: $62.25\%$, cost: $\$47.96$), the \emph{Avengers-Pro} can attain  $7\%$ performance gain with a comparable cost (average accuracy: $66.66$, cost: $\$47.13$), and cut $27\%$ cost with a comparable performance (average accuracy: $62.66$, cost: $\$35.05$). 
By varying the trade-off parameter $\alpha$, the \emph{Avengers-Pro} achieves an even more favorable balance between performance and efficiency. For example, to 
reach 90\% of {GPT-5-medium}'s performance---a level comparable to {Gemini-2.5-pro}---the \emph{Avengers-Pro} 
reduces cost by 63\% relative to GPT-5-medium and by 81\% relative to Gemini-2.5-pro. 
Furthermore, we observe that the \emph{Avengers-Pro} achieves a Pareto frontier: for any fixed cost, it consistently delivers the highest performance among all models at that expenditure. Conversely, for any fixed performance target, it provides the lowest cost compared to other models attaining the same accuracy.

Note that \emph{Avengers-Pro} is not the first study to explore test-time routing. As we will discuss in detail later, there is a growing line of research that leverages router-based methods to harness collective intelligence from multiple models~\cite{routerdc,zhuang2024embedllm,zhang2025avengers}. While most works in this area focus primarily on improving overall performance~\cite{routerdc,zhuang2024embedllm}, a few recent studies have started to investigate the trade-off between performance and efficiency by routing among smaller and larger models~\cite{jitkrittum2025universal,hu2024routerbench}.
Compared to these prior studies, the \emph{Avengers-Pro} stands out for its simplicity and effectiveness. Most previous approaches rely on training additional neural networks and require retraining when incorporating new models. In contrast, \emph{Avengers-Pro} involves only three lightweight operations, requires no neural network training, and is straightforward to implement. Moreover, it is highly reproducible and does not depend on hand-crafted prompts. To incorporate newly available models, it only requires an incremental evaluation of the new models on the dataset.
Despite its simplicity, the \emph{Avengers-Pro} is remarkably effective. To our knowledge, this is the first demonstration that test-time routing can be used to surpass state-of-the-art proprietary single models in terms of both efficiency and performance.

\section{Related Work}

\begin{figure*}[!t]
    \centering \includegraphics[width=1\linewidth]{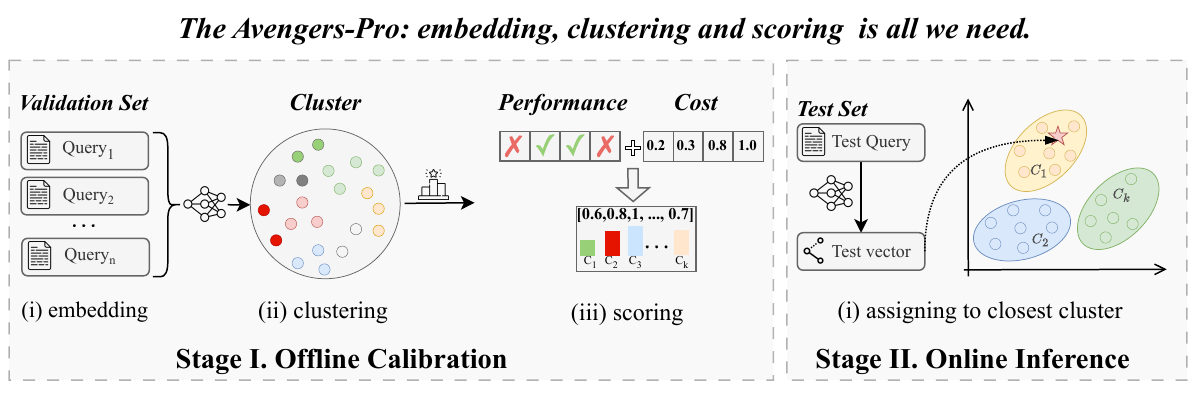}
    \caption{The \emph{Avengers Pro}, a unified framework for dynamic routing, optimizes performance-efficiency trade-offs by intelligently ensembling language models.
    }
    \label{fig:method}
\end{figure*}

The harnessing of collective intelligence from multiple models constitutes one of the frontiers of AI and ML research, and has recently attracted much interest~\citep{lu2024merge,guo2024large,zhang2025if,subramaniam2025multiagent,wan2025rema,zheng2025decouple}. Existing approaches in this area generally fall into three paradigms: router-based, mixture-based, and merging-based methods.
This work is most closely aligned with the router-based paradigm. 
 
The main goal of most router-based methods is to enhance the overall performance of a set of smaller models through query routing; a neural network-based router is often trained to select, for each incoming query, the model most capable of handling it~\citep{chenfrugalgpt,shnitzerlarge,ong2024routellm,feng2025graphrouter,shen2024learning,huang2025routereval}.
LLM-Blender~\citep{jiang2023llm} utilizes pairwise comparisons to select the top-\( k \) models for each query and then fuses their outputs to enhance performance. ZOOTER~\citep{lu2024routing} proposes reward-guided query routing, employing tag-based label enhancement to improve training stability.
RouterDC~\citep{chen2024routerdc}, which employs dual contrastive learning to enhance routing accuracy, and EmbedLLM~\citep{zhuang2024embedllm} leverages learned compact model embeddings along with query embeddings to predict routing correctness. Additionally, Model-SAT~\citep{zhang2025capability} generates capability instructions from model aptitude outcomes and employs text-aligned embeddings to guide a lightweight LLM in selecting optimal candidate models. 
More ecently, few recent studies have also explored routing that account for the trade-off between performance and computational cost. 
Routellm~\citep{ong2024routellm} trains a binary classifier using preference data to dynamically route queries to either a stronger or a weaker LLM during inference.
GraphRouter~\citep{feng2025graphrouter} constructs a heterogeneous graph comprising task, query, and LLM nodes, and predicts the performance-cost score via an edge prediction mechanism.
RouterBench~\citep{hu2024routerbench} introduces a suite of benchmarks. 
Our work is most closely related to \cite{zhang2025avengers} and \citep{jitkrittum2025universal}, which both employ clustering in their routers. While the former focuses only on performance, the latter considers model efficiency to a be universal cost across different tasks. The \emph{Avengers-Pro} is the first study to show that test-time routing can be used to surpass state-of-the-art proprietary single models in terms of both efficiency and performance.

Driven primarily by efficiency, another related line of research explore single-model \emph{hybrid reasoning} methods that trim latency and token cost by adjusting the amount or stages of “thinking” within one model (e.g., Qwen3 “thinking” mode~\footnote{https://huggingface.co/Qwen/Qwen3-235B-A22B} and DeepSeek-V3.1~\footnote{https://huggingface.co/deepseek-ai/DeepSeek-V3.1}), or learn to allocate a reasoning budget via policies\cite{han2025tokenbudgetawarellmreasoning} or test-time guidance\cite{jiang2025think}. Representative approaches include adaptive “think/non-think” modes, hybrid-reasoning training pipelines, and token-budget or inference-aware policies. While such methods can cut redundant CoT tokens or speed up simple cases, they \emph{cannot} realize cross-model complementarity by design (all computation remains within a single backbone), and their effectiveness is often modest~\cite{han2025tokenbudgetawarellmreasoning,kang2025c3ot,alomrani2025reasoning}.

\section{Routing for Performance-Efficiency Trade-off}

The \emph{Avengers-Pro}  ensembles a set of heterogeneous LLMs of varying capabilities and efficiencies with a router.
Appropriate routing depends on an accurate understanding of each model’s capability and efficiency across different types of tasks or queries. To build this understanding, the router requires a set $\mathcal{D}$ of labeled query–answer pairs.
Each query $d \in \mathcal{D}$ is first encoded into a semantic vector using a text \textbf{embedding} model. These embeddings are then grouped into $k$ clusters using a \textbf{clustering} algorithm, producing a set $\mathcal{C} = \{c_1, \dots, c_k\}$, where each cluster represents a semantically coherent query type. 

Let $\mathcal{M}$ denote the set of models in our system.
We evaluate each model $i\in \mathcal{M}$ on $\mathcal{D}$, measures its performance and efficiency within each cluster.
Let  $\mathbf{p}^i = [p_1^i, \dots, p_k^j]^\top$ be a cluster-wise \textbf{performance profile} for model $i$, where $p_j^i$ denotes model $i$’s accuracy on queries within cluster $c_j$.
Similarly, let  $\mathbf{q}^i = [q_1^i, \dots, q_k^j]^\top$ be a cluster-wise \textbf{efficiency profile} for model $i$, where $q_j^i$ denotes model $i$’s efficiency on queries within cluster $c_j$.
We measure the efficiency in terms of cost such that $q_j^i$ denotes the total cost incurred by model $i$ to answer all queries within cluster $c_j$.

We calculate the \textbf{cluster-wise performance-efficiency score} $x_j^i$ for model $i$ on $c_j$  by
$$
    x_j^i = \alpha \, \tilde{p}_j^i + (1 - \alpha) \, (1-\tilde{q}_j^i),
$$
where $\alpha \in [0, 1]$ controls the trade-off between performance and efficiency, and $\tilde{p}_j^i$ and $\tilde{q}_j^i$ are the normalized values of $p_j^i$ and $q_j^i$. The normalization is given by 
$$
    \tilde{p}_{j}^i = \frac{p_{j}^i - p_j^{\min}}{p_{j}^{\max} - p_{j}^{\min}}, \quad
      \tilde{q}_{j}^i = \frac{q_{j}^i - q_j^{\min}}{q_{j}^{\max} - q_{j}^{\min}},
    $$
where $p_j^{\min}$ and $p_j^{\max}$ (or $q_j^{\min}$ and $q_j^{\max}$) denote the minimum and maximum performance (or cost) among all models for cluster $j$.



During inference, an incoming query is encoded  with the  text embedding model, and is assigned to the top-$p$ nearest cluster(s) in the embedding space. For each model $i\in \mathcal{M}$, we sum up its cluster-wise performance-efficiency scores over those top-$p$ clusters. The model with the highest sum of those scores is selected to generate the response.

\section{Experiments}
Our experiments compare the performance and efficiency of \emph{Avengers-Pro} against leading single models.

\subsection{Experimental Settings}
\subsubsection{Models}
We consider 8 leading models, which vary in capability and efficiency, as follows: 
\begin{itemize}[leftmargin=*]
\setlength\itemsep{0em} 
    \item \textbf{Google}: Gemini-2.5-flash~\cite{comanici2025gemini}, Gemini-2.5-Pro~\cite{comanici2025gemini}.
    \item \textbf{Anthropic}: Claude-4.1-opus~\cite{2025claude}, Claude-4-sonnet~\cite{sonnet}.
    \item \textbf{OpenAI}: GPT-5-chat~\cite{2025gpt5}\footnote{\href{https://platform.openai.com/docs/models/gpt-5-chat-latest}{\textbf{GPT-5-chat}}: points to the GPT-5 snapshot currently used in ChatGPT; OpenAI recommends \texttt{gpt-5} for most API usage, while \texttt{gpt-5-chat} exposes the latest improvements for chat use cases. Do not support function/tool call now.
}, GPT-5-medium~\cite{2025gpt5}\footnote{\href{https://platform.openai.com/docs/models/gpt-5}{\textbf{GPT-5-medium}}: GPT-5 is OpenAI's flagship model for coding, reasoning, and agentic tasks across domains. GPT-5-medium denotes GPT-5 with \texttt{reasoning\_effort=medium}.}.
    \item \textbf{Qwen}: Qwen3-235B-A22B-2507 (or Qwen3)~\cite{yang2025qwen3}, Qwen3-235B-A22B-thinking-2507 (or Qwen3-thinking)~\cite{yang2025qwen3}.
\end{itemize} 
We access these models through the OpenRouter API\footnote{https://openrouter.ai/},  as its standardized interface simplifies the process of running identical experiments across multiple models. The pricing for these models is detailed in Table 1. 
Prices for the Qwen3 family may vary across providers; throughout this paper we report the prices listed by OpenRouter.

\begin{table}[!htb] 

        \centering 
        \caption{Model cost information~(OpenRouter).} 
        \label{tab:price}
        {%
\small
\begin{tabular}{@{}lrr@{}}
\toprule
\textbf{Model} & \textbf{Input Price} & \textbf{Output Price} \\ 
&(\$/1M tokens)&(\$/1M tokens)\\
\midrule
\textbf{Gemini-2.5-flash} & 0.30 & 2.50 \\
\textbf{Gemini-2.5-Pro} & 1.25 & 10 \\
\textbf{Claude-4.1-opus} & 15 & 75 \\
\textbf{Claude-4-sonnet} & 3 & 15 \\
\textbf{GPT-5-chat} & 1.25 & 10 \\
\textbf{GPT-5-medium} & 1.25 & 10 \\
\textbf{Qwen3-235B-A22B-25074} & $\approx$0.13 & $\approx$0.6 \\
\textbf{Qwen3-235B-A22B-thinking-2507} & $\approx$0.13 & $\approx$0.6 \\

\bottomrule
\end{tabular}%

    }
\end{table}

\subsubsection{Benchmarks}
We consider 6 challenging benchmarks, as summarized in Table~\ref{tab:benchmarks}, covering advanced reasoning and general knowledge:

\begin{table}[!htb] 
        \centering 
        \caption{ Benchmark information.} 
        \label{tab:benchmarks}
\begin{tabular}{@{}llr@{}}
\toprule

\textbf{Dataset} & \textbf{Metrics} & \textbf{Size} \\ \midrule
ARC-AGI-v1~\cite{chollet2024arc} & pass@1 & 200 \\
GPQA-Diamond~\cite{rein2024gpqa} & pass@1 & 198 \\
HLE~\cite{phan2025humanity} & pass@1 & 500 \\
LiveCodeBench-v6~\cite{jain2024livecodebench} & pass@1 & 1,055 \\
$\tau^2$-bench~\cite{barres2025tau} & pass@1 & 150 \\
SimpleQA~\cite{wei2024measuring} & pass@1 & 500 \\
\midrule
Total & & 2,603 \\
\bottomrule
\end{tabular}%

\end{table}

\begin{itemize}[leftmargin=*]
\setlength\itemsep{0em} 
    \item \textbf{GPQA-Diamond}~\cite{rein2024gpqa}: A graduate-level google-proof Q\&A benchmark.
    \item \textbf{Human's Last Exam (HLE)}~\cite{phan2025humanity}: A frontier multi-modal benchmark of closed-ended academic questions. In this study, we use the \emph{text-only} setting without custom patches, tool use, or retrieval during evaluation. For efficiency and reproducibility, we use the first \textbf{500} questions from the released pool and report accuracy under the official evaluation protocol.
    \item \textbf{ARC-AGI}~\cite{chollet2024arc}: A benchmark focused on fluid intelligence, testing the ability to reason and solve novel problems. We use the first \textbf{200} questions from the released pool and report accuracy under the official evaluation protocol.
    \item \textbf{SimpleQA}~\cite{wei2024measuring}: A factuality benchmark for short, fact-seeking questions. We use the \emph{official} implementation with the default configuration and report accuracy under the official scoring. We evaluate on a subset of \textbf{500} examples uniformly sampled from the released dataset.
    \item \textbf{LiveCodeBench}~\cite{jain2024livecodebench}: A dynamic, contamination-controlled coding benchmark that continuously ingests newly released problems. We evaluate on the latest public release (\textbf{v6}) using the \emph{official} implementation and evaluation harness with the default configuration, without custom patches or post-processing.
    \item \textbf{$\tau^2$-bench}~\cite{barres2025tau}: A controlled testbed for agents that must reason effectively and guide user actions, we randomly sampled 50 examples from each of the three categories.
\end{itemize}

For all benchmarks, we use the official implementations with their recommended prompts and hyperparameters; for context length, we set each model to its documented maximum context window and apply no additional truncation, chunking, or sliding-window processing.

\begin{table*}[!h]
\small
\centering
\caption{The performance and efficiency of \emph{Avengers-Pro} vs. single models. 
Note that GPT-5-chat has no score on the $\tau^2$-bench benchmark because this model does not support tool calling.
\textbf{Bold} indicates the best performance of a given benchmark, and \underline{underline} indicates the second-best. 
With $\alpha=0.1$, \emph{Avengers-Pro},  surpasses GPT-5-medium in average accuracy with a $7\%$ performance gain. 
With $\alpha=0.53$, it matches  GPT-5-medium's average accuracy, while cutting the cost by $27\%$.
With $\alpha=0.39$, it reaches 90\% of GPT-5-medium's performance at a $63\%$  lower cost.
}
\label{tab:main-result}
\resizebox{\textwidth}{!}{
\begin{tabular}{@{}lcccccccr@{}}
\toprule
\textbf{Setting}         
& \textbf{ARC-AGI} 
& \textbf{GPQA-Diamond}
& \textbf{HLE}
& \textbf{LiveCodeBench}
& \textbf{SimpleQA}
& \textbf{$\tau^2$-bench}
& \textbf{Avg. A}  
& \textbf{Cost}
\\ \midrule
\textbf{Gemini-2.5-flash}  & 9.62 & 21.72  & 7.20 & 62.84 & 28.99 & 36.67 & 27.84 & \$7.10   \\
\textbf{Gemini-2.5-pro} & 33.08 & 84.85 & 23.09 & 78.67 & 54.80 & 62.00 &  56.08   & \$94.87 \\
\textbf{Claude-4.1-opus} & 22.12 & 74.24  & 6.41 & 64.07 &  31.00 & 74.00 &   45.31 &  \$117.40    \\
\textbf{Claude-4-sonnet}  & 16.15 & 68.69  & 4.60 & 59.05 & 15.00 & 64.00 & 37.92 & \$25.35  \\
\textbf{Qwen3}    & 9.22 & 58.59 & 9.22 & 66.26 & 53.00 & 53.33 & 41.60 & \$2.73   \\
\textbf{Qwen3-thinking}  & 19.23 & 80.81  & 12.68 & 77.99 & 44.60 & 53.33 &  48.11 & \$13.99  \\
\textbf{GPT-5-chat}   & 6.73 & 73.73  & 7.80 & 63.60 & 40.20 & - & 38.41 & \$4.04 \\
\textbf{GPT-5-medium}  & 44.42 & \underline{84.85}  & 26.20 & 88.44 & 47.60 & \textbf{82.00} & 62.25 & \$47.96  \\
\midrule
\bfseries \emph{Avengers Pro} ($\alpha=0$)  
& 15.33 & 58.67 & 10.13 & 66.94 & 46.27 & 0.00 & 32.89 & \$1.08 \\ 
\bfseries \emph{Avengers Pro} ($\alpha=0.25$)$^1$ 
& 29.33 & 67.00 & 10.00 & 76.53 & 53.60 & 72.89 & 51.56 & \$9.69  \\ 
\bfseries \emph{Avengers Pro} ($\alpha=0.39$)$^2$  
& 29.33 & 78.67 & 12.67 & 84.79 & 55.07 & 76.89 & 56.24 & \$17.81 \\ 
\bfseries \emph{Avengers Pro} ($\alpha=0.53$)$^3$  
& \underline{51.67} & 80.00 & 25.46 & 87.45 & 54.93 & 76.44 & 62.66 & \$35.05 \\ 
\bfseries \emph{Avengers Pro} ($\alpha=0.8$)
& \textbf{59.67} 
& 81.00 & \underline{27.60} & \underline{89.34} & \textbf{56.93} & 78.22 & \underline{65.46} & \$44.65 \\ 
\bfseries \emph{Avengers Pro} ($\alpha=1$) 
& \textbf{59.67} & \textbf{85.67}  & \textbf{28.67} & \textbf{89.59} & \underline{56.40} & \underline{80.00} & \textbf{66.66} & \$47.13 \\ 
\bottomrule
\end{tabular}
}
\end{table*}

\subsubsection{Implementation Details}
We use k-means clustering with $k=60$ clusters. Each query is encoded by the \textit{Qwen3-embedding-8B}~\cite{zhang2025qwen3embeddingadvancingtext} into a 4,096-dimensional semantic vector. Following common practice in routing~\cite{routerdc, zhuang2024embedllmlearningcompactrepresentations, zhang2025avengers}, we randomly split the data: 70\% is used to fit the clustering model and estimate per-cluster statistics, and the remaining 30\% is reserved for routing and evaluation. At inference time, we compute the embedding of the incoming query and retrieve the top-$p$ nearest clusters ($p=4$) in the embedding space. For each model $i$, we then sum its cluster-wise cost–capability scores $q_j^i$ over these three clusters and select the model with the highest total to generate the response.

\subsection{Results and Analysis}
\begin{figure*}[tb!]
    \centering
    \includegraphics[width=1\linewidth]{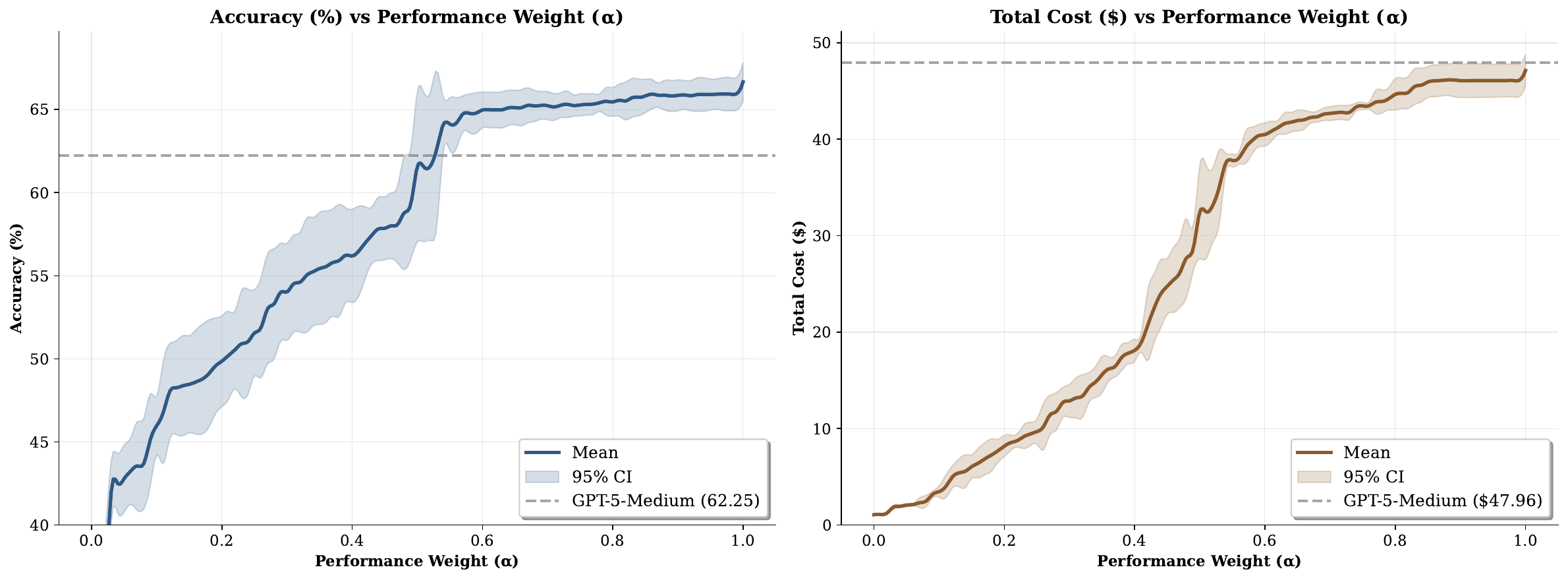}
    \caption{Effects of the trade-off parameter $\alpha$ on the performance and efficiency. A greater value of $\alpha$ prioritizes performance over efficiency. The increase in performance is usually accompanied the increase in cost.  }
    \label{fig:alpha-curves}
\end{figure*}

We present the comparisons of \emph{Avengers-Pro} and single models in terms of performance and efficiency in Table 3. We show how the trade-off parameter $\alpha$ affects the performance and efficiency in Figure~\ref{fig:alpha-curves}. 
We show the proportion of model usage by \emph{Avengers-Pro} in Figure~\ref{fig:model_distribution}.

\subsubsection{\textit{Avengers-Pro} outperforms top single models.} 
Of the eight single models evaluated, GPT-5-medium demonstrates the highest average accuracy ($62.25\%$) across the six benchmarks. This is followed by Gemini-2.5-pro ($56.08\%$) and Qwen3-thinking ($48.11\%$), respectively. The \emph{Avengers-Pro} surpasses the performance of \emph{all} individual models with a sufficiently large value of $\alpha$, prioritizing performance over efficiency. 
Specifically, its average accuracy is up to $66.66\%$ with $\alpha=1.0$, which is $7\%$ higher compared to GPT-5-medium and $19\%$ higher compared to Gemini-2.5-pro. 

\subsubsection{\textit{Avengers-Pro} achieves a superior performance-efficiency trade-off.}
At a performance level comparable to the strongest single model GPT-5-medium, \emph{Avengers-Pro} ($\alpha=0.53$) incurs significantly lower costs, resulting in a cost reduction of
$27\%$.
Similarly, at a $90\%$ performance level of GPT-5-medium, the \emph{Avengers-Pro} ($\alpha=0.39$) cuts cost by $63\%$.
At a performance level comparable to the second-strongest single model Gemini-2.5-pro, it ($\alpha=0.39$) reduces cost by $81\%$.
At a performance level comparable to Cluade-4.1-opus, it ($\alpha=0.25$) achieves a cost reduction of $92\%$.
Moreover, as shown in Figure~\ref{fig:big-picture}A, the \emph{Avengers-Pro} achieves a Pareto frontier---no single model can simultaneously deliver higher performance and greater efficiency than \emph{Avengers-Pro}. In other words, \emph{Avengers-Pro} offers the highest performance for any given cost and the lowest cost for any given level of performance.

\begin{figure*}[!t]
    \centering
    \includegraphics[width=1\linewidth]{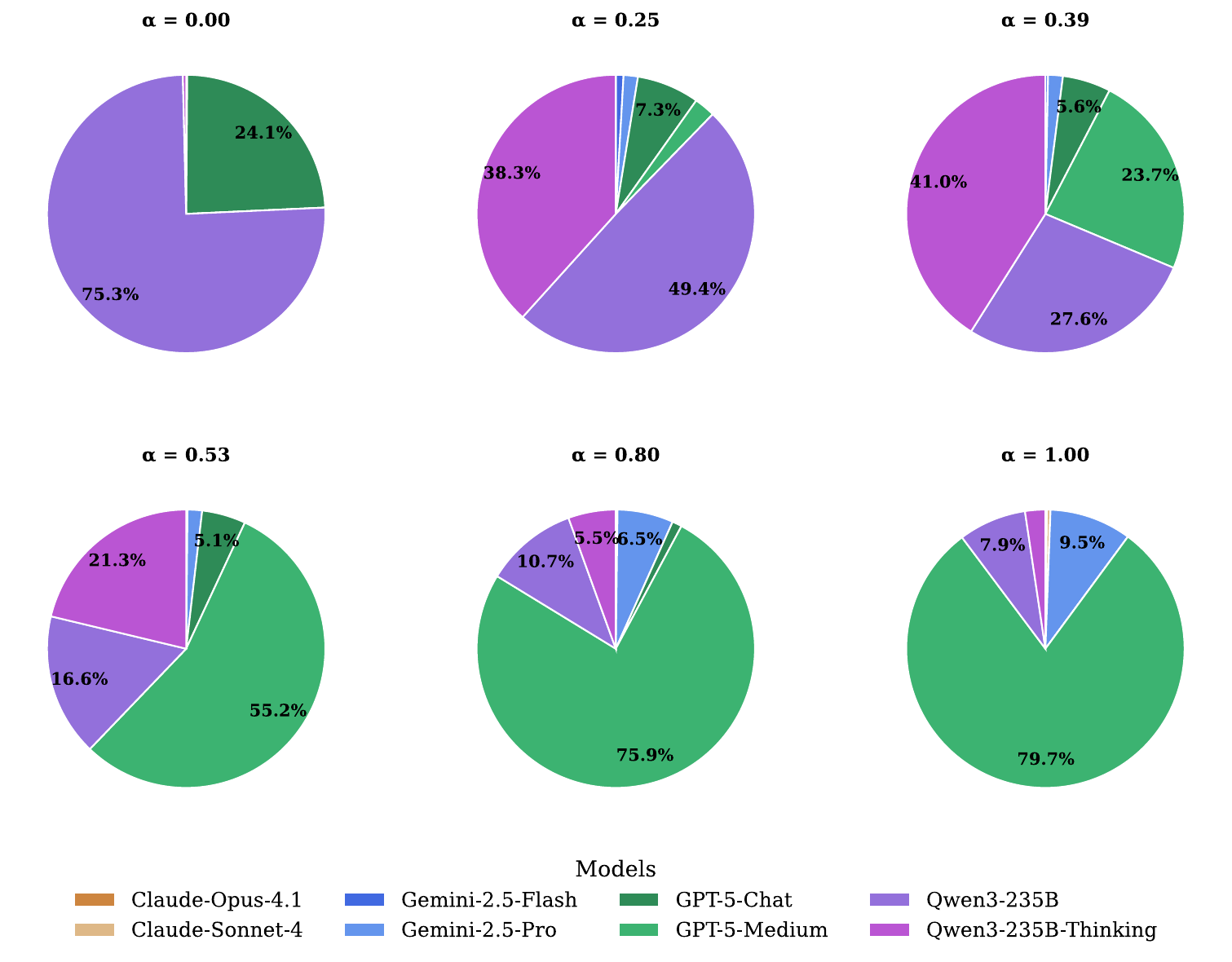}
    \caption{Proportion of model usage, given various trade-off parameters $\alpha$. When $\alpha$ is low, \emph{Avengers-Pro} tend to route queries to Qwen3 and Qwen3-thinking. With a greater value of $\alpha$, \emph{Avengers-Pro} favors GPT5-medium and Qwen3-thinking.}
    \label{fig:model_distribution}
\end{figure*}

\subsubsection{Effects of the trade-off parameter} 
As shown in Figure~\ref{fig:alpha-curves}, we gradually increase the trade-off parameter $\alpha$, placing more weight on performance. Accuracy (left) climbs quickly for small $\alpha$ and then saturates around $\alpha\!\approx\!0.6$; the shaded 95\% CI also narrows as $\alpha$ grows, indicating more stable outcomes once the router consistently calls stronger models. Cost (right) stays near its minimum up to $\alpha\!\approx\!0.4$, then rises steeply before tapering off. The gray dashed baselines mark GPT-5-medium (62.25\% accuracy; \$47.96 cost): our curve exceeds the accuracy baseline already at moderate $\alpha$ while remaining below the cost baseline across a broad range before very high $\alpha$. Taken together, the curves reveal two elbows---around $\alpha\!\approx\!0.4$ and $\alpha\!\approx\!0.6$---that correspond to regime changes in routing: (i) \emph{efficiency-first} ($\alpha\!\lesssim\!0.4$), where the router mostly selects cheaper models and yields low cost with moderate accuracy; (ii) a \emph{transition} band ($0.4\!<\!\alpha\!<\!0.6$), where accuracy improves rapidly per marginal dollar as the router begins invoking stronger models on harder clusters; and (iii) \emph{performance-first} ($\alpha\!\gtrsim\!0.6$), where accuracy is near its ceiling and additional $\alpha$ mainly buys cost increases.

\subsubsection{Proportion of model usage} 
As shown in Figure~\ref{fig:model_distribution}, when $\alpha$ is low, \emph{Avengers-Pro} tends to favor the Qwen3 and Qwen3-thinking model, routing a great proportion of queries to these two models with a low unit price. As $\alpha$ increases, the usage of GPT-5-medium rises rapidly; concurrently, the usage of Gemini-2.5-pro and Claude-opus-4.1, which excel at complex reasoning but have a higher unit price, also increases.
Consistent with Figure~\ref{fig:big-picture}, model usage correlates with proximity to \emph{Avengers-Pro}’s Pareto frontier: models closer to the frontier (GPT-series, Qwen3-series) see higher utilization, whereas those farther away (claude-series, gemini-series) are selected less frequently for a given $\alpha$.

\section{Conclusions}
In this work, we introduce \emph{Avengers-Pro}, a 
test-time routing framework integrating different LLMs to optimize the trade-off between performance and efficiency. By dynamically selecting exactly one model for each incoming query, \emph{Avengers-Pro} optimizes both cost and accuracy. Our experiments involving 8 leading LLMs and 6 challenging benchmarks demonstrate that \emph{Avengers-Pro} can surpass the strongest single model, GPT-5-medium, by up to 7\% in accuracy and match its performance at a 27\% lower expense. Moreover, \emph{Avengers-Pro} achieves a Pareto frontier, consistently delivering the best performance on any given budget and the lowest cost given any performance target. Our results highlight the significant potential of an intelligent test-time routing framework in creating more powerful, efficient, and scalable LLM systems. 

\begin{acks}
This work was supported by a locally commissioned task from the Shanghai Municipal Government. The authors thank Dr. Qiaosheng Zhang for helpful discussions, and collaborators who contributed meaningfully to our earlier project the \emph{Avengers}~\cite{zhang2025avengers}. This work was supported by the National Natural Science Foundation of China (Grant No. 6250076060).
\end{acks}


\bibliographystyle{ACM-Reference-Format}
 \bibliography{ref}

\end{document}